\documentclass[conference]{IEEEtran}
\IEEEoverridecommandlockouts

\usepackage{cite}
\usepackage{amsmath,amssymb,amsfonts}
\usepackage{algorithmic}
\usepackage{booktabs}
\usepackage{graphicx}
\usepackage{textcomp}
\usepackage{array}

\usepackage{hyperref}

\usepackage{xcolor}
\def\BibTeX{{\rm B\kern-.05em{\sc i\kern-.025em b}\kern-.08em
    T\kern-.1667em\lower.7ex\hbox{E}\kern-.125emX}}
\begin{document}

\title{Semantic to Structure: Learning Structural Representations for Infringement Detection\\
}

\author{\IEEEauthorblockN{
Chuanwei Huang\IEEEauthorrefmark{2},
Zexi Jia\IEEEauthorrefmark{2}, 
Hongyan Fei\IEEEauthorrefmark{2}, 
Yeshuang Zhu, 
Zhiqiang Yuan, 
Jinchao Zhang*, 
Jie Zhou* }
\IEEEauthorblockA{Peking University, Beijing, China \\  Wechat AI, Tencent, China}
}

\maketitle

\begin{abstract}
Structural information in images is crucial for aesthetic assessment, and it is widely recognized in the artistic field that imitating the structure of other works significantly infringes on creators' rights. The advancement of diffusion models has led to AI-generated content imitating artists' structural creations, yet effective detection methods are still lacking. In this paper, we define this phenomenon as "structural infringement" and propose a corresponding detection method. Additionally, we develop quantitative metrics and create manually annotated datasets for evaluation: the SIA dataset of synthesized data, and the SIR dataset of real data. Due to the current lack of datasets for structural infringement detection, we propose a new data synthesis strategy based on diffusion models and LLM, successfully training a structural infringement detection model. Experimental results show that our method can successfully detect structural infringements and achieve notable improvements on annotated test sets.
\end{abstract}

\begin{IEEEkeywords}
Image Infringement, Diffusion Models, Contrastive Learning
\end{IEEEkeywords}
\section{Introduction}




The structural information of images has long been an important aspect of aesthetic assessment~\cite{zhang2021image}. Image structure encompasses the geometric structure and the arrangement of visual elements with images, making it a critical aspect of artistic and commercial creations. Therefore, many artistic works mimic the structural composition of other works rather than directly imitating their content to avoid detection. On the other hand, the rapid development of diffusion models~\cite{ho2020denoising,podell2023sdxl} enables the synthesis of high-quality images, but it also exacerbates the issue of image infringement.
Previous studies~\cite{somepalli2023understanding, somepalli2023diffusion, he2024fantastic, wang2024evaluating} have highlighted that images produced by diffusion models may infringe on existing works from different perspectives, including both semantic and structural. These cases of image infringement greatly damage the copyright of creators, necessitating the detection of image infringement.

\begin{figure*}[htbp]
\centerline{\includegraphics[width=0.85\textwidth]{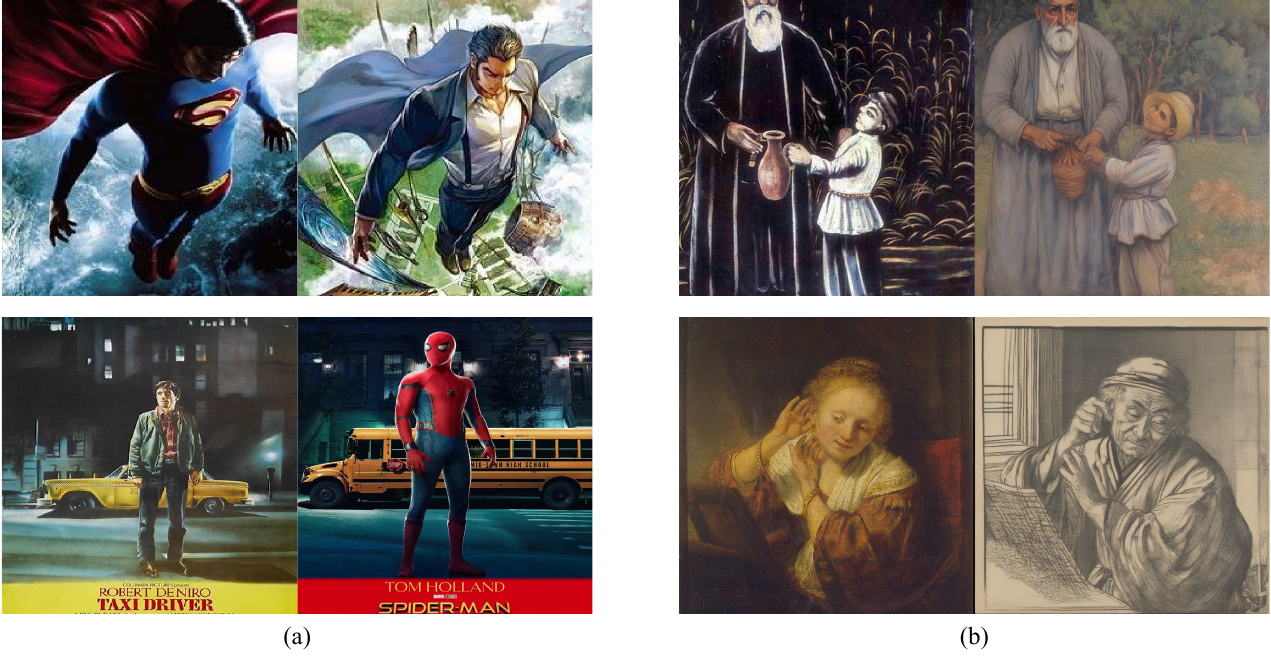}}
\caption{Structural infringement image pairs in SIR and SIA datasets. (a): The SIR dataset encompasses image pairs that exhibit structural infringement in the real world. (b) The SIA dataset includes image pairs with structural infringement generated by diffusion models, with real images on the left and synthetic images on the right. Despite the low content similarity, these pairs exhibit high structural similarity, indicating potential structural infringement.}
\label{fig}
\end{figure*}

Commonly-used infringement detection methods predominantly focus on semantic infringement. 
These detection methods often overlook cases where images exhibit high structural similarity but low semantic similarity, as illustrated in Fig.~\ref{fig}. In this work, we refer to this phenomenon as \textit{structural infringement}.
To efficiently detect structural infringement, we introduce a novel type of image representation, termed the \textit{Image Structural Representation}, which describes the geometric and positional information within images in a fine-grained manner. 

Traditional image representation learning methods primarily utilize self-supervised learning methods for image representation learning. Methods like MoCo \cite{chen2021empirical} and DINO \cite{oquab2023dinov2} \cite{caron2021emerging} leverage different data augmentations to create multiple views of the same image.
By maximizing the agreement between different views, the model learns invariance to these data augmentations, thereby extracting rich semantic information.
Although these methods are effective at extracting rich semantic information, they struggle to extract meaningful structural information. This is due to the current lack of datasets that possess only similar structures, preventing the extraction of structural information. 
To tackle the issue of data scarcity, we propose a novel data synthesis pipeline that generates image pairs with similar structural information but different semantic content. These synthesized pairs are then used to train the image structural representation extractor, achieving a comprehensive understanding of images and providing a robust framework for structural infringement detection.

Previous work most closely related to our image structural representation is image layout representation~\cite{zhao2024self, 9577542, hou2020object}, which converts image layouts into concise vectors to support various downstream tasks such as image layout classification and retrieval. 
In~\cite{zhao2024self}, primitives in the image are used to construct a heterogeneous graph and subsequently learn the image layout representation in a self-supervised manner. 
However, layout primarily describes the coarse-grained arrangement of elements within images, whereas structural infringement detection requires finer-grained structural information, including geometric and positional information.

In this work, we extend the concept of structural combination information in aesthetic assessment to AIGC images, and we decouple structural information from semantic information. We define \textit{structural infringement} as the occurrence where one image, whether human-created or AI-generated, infringes upon the structural information of another image.
To efficiently detect structural infringement, we propose to train a model dedicated to extracting structural information from images, which we term as image structural representation.
The primary contributions of this work are as follows:
\begin{itemize}
    \item We analyze the phenomenon of infringement in realistic and synthetic images from the perspective of structural information and define this phenomenon as structural infringement.
    \item We propose to extract structural compositional information from images and design a novel data synthesis strategy for learning image structural representations.
    \item To evaluate the capabilities of different methods, we construct two manually annotated structural infringement test sets, SIA and SIR dataset. Our proposed method achieved state-of-the-art results on these datasets.
\end{itemize}
The SIA and SIR datasets are available at: \href{https://drive.google.com/drive/folders/1jQNyt5IEP2Lx5G0APCZDbK8vCLB_oijX?usp=sharing}{\textit{dataset link}}.
\section{Methods}
\begin{figure*}[htbp]
\centerline{\includegraphics[width=0.9\textwidth]{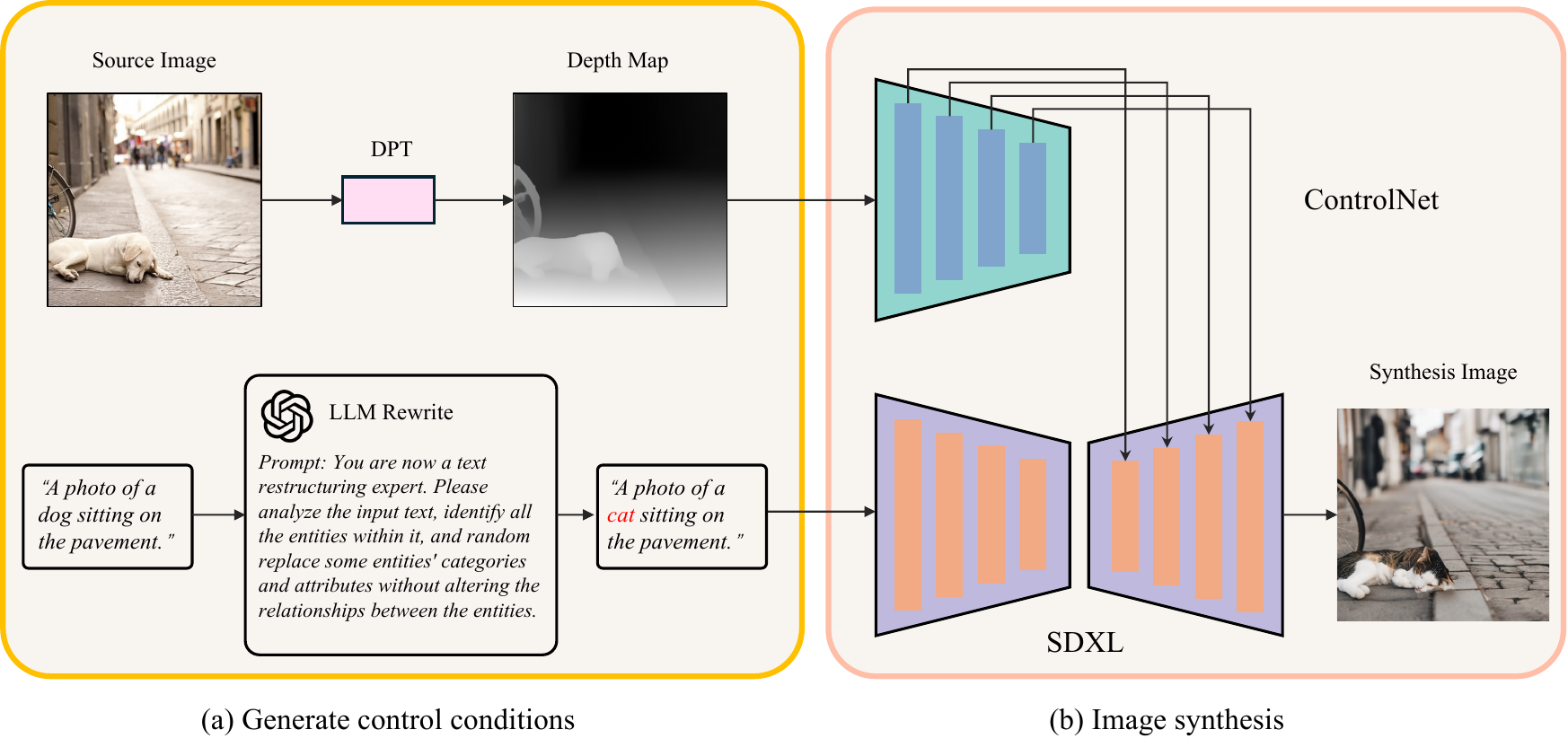}}
\caption{Data synthesis pipeline. (a) Given a source image with a caption description, the depth map is first extracted using DPT to capture the main structural information. Subsequently, the LLM is used to modify the attributes of the main objects in the caption to change the semantic information. (b) The text and image condition are then input into SDXL and ControlNet respectively, generating images with high structural similarity but low semantic similarity to the source image.}
\label{fig:data}
\end{figure*}

Due to the current lack of research on learning image structural information, there is a shortage of relevant training data and corresponding benchmarks.
To address this issue, we propose a novel data synthesis pipeline to generate image pairs that have high structural similarity but low semantic similarity, as illustrated in Fig.~\ref{fig:data}. Base on this, we train a image structural representation extractor.

With the development of diffusion models, it becomes possible to generate photo-realistic images and accept various conditions.
ControlNet~\cite{zhang2023adding} integrates guidance signals into diffusion models, enabling more precise control over the generative process and supporting various condition signals. Consequently, we choose to use "SDXL~\cite{podell2023sdxl} + ControlNet" to generate images that retain structures similar to the source image.
Given a source image $x_{\text{src}}$, we use the DPT~\cite{Ranftl_2021_ICCV} model to generate its depth map, which is then used as a control condition for ControlNet to ensure the generated image $x_{\text{syn}}$ has a similar structure. The depth map is used as the control condition because it preserves the rough geometric and positional information of the main elements within the image. Other control conditions, such as Canny edges, may include excessively detailed information, resulting in a generation quality decrease.

However, a potential issue with the aforementioned approach is that the generated images are not only structurally similar to the source image but also semantically similar. This high degree of semantic similarity is not conducive to our subsequent training of the image structural representations. To address this, we employ a Large Language Model (LLM) to rewrite the caption corresponding to the source image. By altering the categories and attributes of the main subjects in the image, we aim to reduce the semantic similarity between the generated and source images. Furthermore, we optionally modify the style in the caption to change stylistic information. 

Training a structural representation extractor from scratch necessitates a substantial amount of labeled data and significant training time. Therefore, we opt to fine-tune pre-trained visual models to enhance their focus on the structural information.
More specifically, we fine-tune a pretrained image encoder using contrastive learning~\cite{chen2020simple, he2020momentum}. 
Contrastive learning has proven to be a powerful technique for self-supervised representation learning. The core idea involves generating two different views of the same image through various data augmentation. These views are then used to train the model such that the representations of the same image are as similar as possible, while the features from different images are pushed apart. 
In this way, the model successfully learns features invariant to various data augmentations, thereby extracting effective semantic information.

In our task, given a pair of images $[x^{\text{src}}_i, x^{\text{syn}}_i]$, generate \(x^{\text{syn}}_i\) from \(x^{\text{src}}_i\) using a diffusion model can be regarded as a complex off-line data augmentation retaining only structural information. 
Following the training paradigm of MoCo, we first input the two images into extractor $E$ and momentum extractor $E'$, obtaining l2 normalized features $\mathbf{f_i}$ and $\mathbf{f_i}'$ respectively. Unlike MoCo, which requires computing the loss after mapping through a predictor, we directly calculate the loss using the features obtained from the extractor. This approach enhances the performance in subsequent retrieval for structural infringement.
We use the InfoNCE loss to maximize the consistency between $\mathbf{f_i}$ and $\mathbf{f_i}'$, and minimize it between $\mathbf{f_i}$ and $\mathbf{f_j}'$, thereby constraining the network to extract only the structural representation. The loss is calculated as follows:
$$  \mathcal{L}_{\text{InfoNCE}} = -\log \frac{\exp(\mathbf{f_i} \cdot \mathbf{f_i}' / \tau)}{\exp(\mathbf{f_i} \cdot \mathbf{f_i}' / \tau) + \sum_{j} \exp(\mathbf{f_i} \cdot \mathbf{f_j}' / \tau)},$$
where the temperature parameter $\tau$ is set to 0.2.


\section{Experiments}

\subsection{Datasets}
\textbf{Testset}: We construct two test datasets to evaluate the model's performance on structural infringement tasks: the Structural Infringement of Artworks (SIA) Dataset and the Structural Infringement of Real Images (SIR) Dataset.

\textit{SIA Dataset} is constructed using a synthetic approach. Initially, 2,000 art images of various styles are randomly selected from WikiArt. Subsequently, we employ the data synthesis pipeline in Fig.~\ref{fig:data} to generate infringing images. Since determining structural infringement ultimately requires subjective human evaluation, each pair of data is manually rated on a scale of 1 to 5, with higher scores indicating a greater degree of infringement. Pairs with average scores greater than 4 are retained, resulting in a testset of 513 pairs.

\textit{SIR Dataset} comprises images manually collected from real-world cases of alleged structural copyright infringement, encompassing various artistic styles such as photographs, comics, and posters. These images undergo a manual filtering process similar to SIA dataset, resulting in the retention of 30 image pairs. Examples from the test sets are shown in Fig.~\ref{fig}.

\textbf{Trainset}: We select 10,000 detailed-caption images from the COCO~\cite{lin2014microsoft} 2017 trainset to generate synthetic training data. GPT-4o rewrites the captions as text conditions for SDXL. We choose to use realistic data instead of artistic works to prevent overfitting due to excessive similarity with the testset.

\begin{figure*}[htbp]
\centerline{\includegraphics[width=0.95\textwidth]{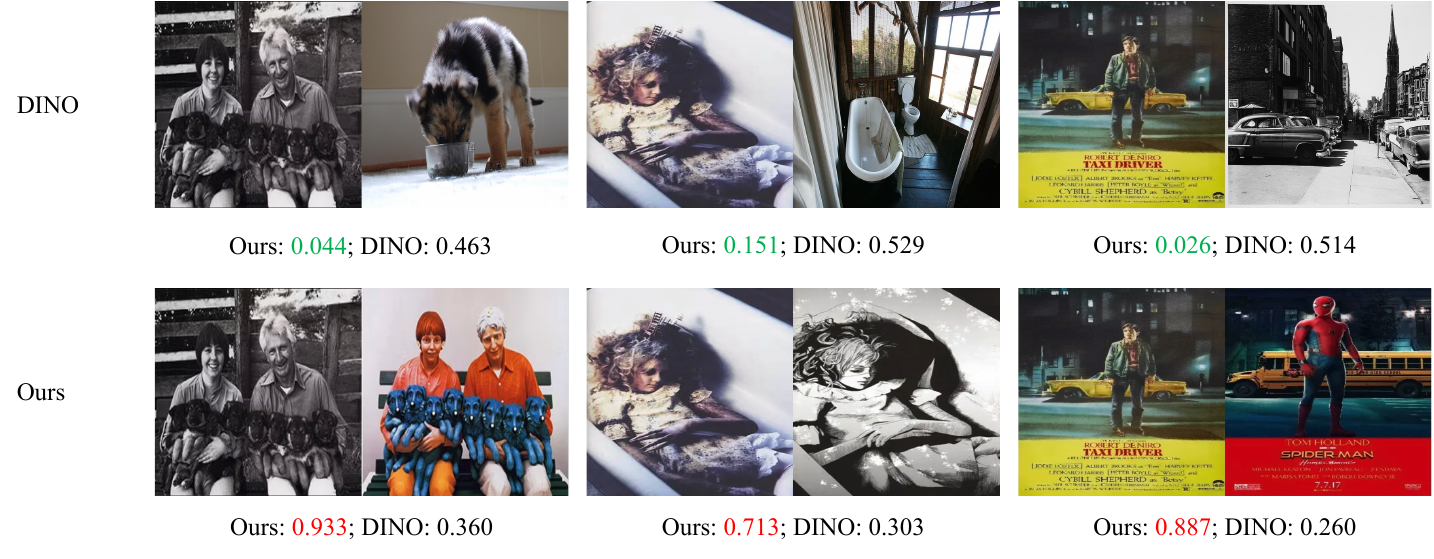}}
\caption{Top-1 retrieval image on SIA datasets using DINO and our proposed image structural representation. For each pair, the left image is the query, and the right image is the retrieval result. The cosine similarity scores for each pair are shown below.}
\label{fig:sim}
\end{figure*}
\subsection{Implementation Details}
We use the pretrained ViT-L~\cite{dosovitskiy2020image} from DINOv2 as the backbone and use LoRA~\cite{hu2021lora} to fine-tune the model to improve training efficiency, with the hyperparameter \( r = 3 \). We set the initial learning rate to \( 1 \times 10^{-4} \) and utilize a cosine learning rate decay schedule. The AdamW~\cite{loshchilov2017decoupled} optimizer is employed for training the model. 
We utilize the Faiss~\cite{johnson2019billion} library to conduct the k-nearest neighbor search for testing.

\subsection{Evaluation}
We evaluate the model's ability to detect structural infringement following the image copy detection~\cite{pizzi2022self} paradigm. Given a query image, we retrieve the most likely infringing reference image from the candidate gallery, resulting in a list of \textbf{\{query, reference\}} pairs with confidence scores. We generate precision-recall curves by adjusting the confidence threshold and use average precision ($\mu AP$) to assess overall performance, similar to instance recognition~\cite{perronnin2009family}. The calculation is as follows:
$$\mu AP = \sum_{i=1}^{N} p(i) \Delta r(i),$$
where \( p(i) \) is the precision at position \( i \) of the sorted precision-recall list, \( \Delta r(i) \) is the difference between the $i$th and $(i-1)$th recall, and \( N \) is the total number of returned predictions. 

We also use features extracted from different methods for image retrieval, visualize their top-1 retrieval results, and report the mAP metrics.

\subsection{Qualitative Results}
To demonstrate that our proposed method primarily extracts structural information from images, we perform image retrieval on the SIA dataset using our learned structural representation and DINO feature. As shown in Fig.~\ref{fig:sim}, the images retrieved using DINO often contain objects of the same category as the query image (e.g., dogs, bathtubs, and vehicles), even though they are not structurally consistent.
In contrast, our proposed structural representation primarily extracts structural information from images. It yields high similarity scores for images with similar structures, whereas images with dissimilar structures will have low similarity scores even if they contain the same semantic concepts.

\subsection{Quantitative Results}
We compare the performance of our proposed image structural representation against other classical image representation learning methods for structural infringement detection DINOv2 and MoCoV3, and the traditional image copy detection method SSCD.
All methods are evaluated with ViT-L as the backbone, except for SSCD, which was tested with both ResNet50~\cite{he2016deep} and ResNeXt101~\cite{xie2017aggregated}.
We compare their performance on the SIA and SIR testsets in Table~\ref{tab:sia} and Table~\ref{tab:sir}.
Due to limited data in the SIR dataset, we add 20,000 images to expand the retrieval gallery. Results show our method significantly outperforms others in detecting structural infringement. 
The performance rankings of different methods on the SIA dataset are roughly consistent with those on the SIR dataset, indicating that our data synthesis pipeline can partially reflect real-world structural infringement phenomena. 



\begin{table}
\centering
\caption{Performance comparison on SIA dataset}
\label{tab:sia}
\begin{tabular}{>{\centering\hspace{0pt}}m{0.15\linewidth}|>{\centering\hspace{0pt}}m{0.12\linewidth}|>{\centering\hspace{0pt}}m{0.12\linewidth}|>{\centering\hspace{0pt}}m{0.12\linewidth}|>{\centering\arraybackslash\hspace{0pt}}m{0.16\linewidth}} 
\hline
         & $\mu AP$            & mAP@1 & mAP@5 & mAP@10  \\ 
\hline
SSCD-50  & 0.102          & 0.275  & 0.328    & 0.339  \\
SSCD-101 & 0.110           & 0.279  & 0.345    & 0.355   \\
DINOv2~  & 0.129          & 0.415 & 0.496    & 0.510 \\
MoCoV3   & 0.120           & 0.359 & 0.434    & 0.445  \\
Ours     & \textbf{0.365} & \textbf{0.667} & \textbf{0.734}    & \textbf{0.743}  \\
\hline
\end{tabular}
\end{table}

\begin{table}
\centering
\caption{Performance comparison on SIR dataset}
\label{tab:sir}
\begin{tabular}{>{\centering\hspace{0pt}}m{0.15\linewidth}|>{\centering\hspace{0pt}}m{0.12\linewidth}|>{\centering\hspace{0pt}}m{0.12\linewidth}|>{\centering\hspace{0pt}}m{0.12\linewidth}|>{\centering\arraybackslash\hspace{0pt}}m{0.16\linewidth}} 
\hline
         & $\mu AP$            & mAP@1 & mAP@5 & mAP@10  \\ 
\hline
SSCD-50  & 0.376         & 0.400  & 0.462    & 0.466  \\
SSCD-101 & 0.450           & 0.467  & 0.511    & 0.515   \\
DINOv2~  & 0.461          & 0.533 & 0.615    & 0.632  \\
MoCoV3   & 0.496          & 0.567 & 0.646    & 0.646 \\
Ours     & \textbf{0.527} & \textbf{0.633} & \textbf{0.667}   & \textbf{0.671}  \\
\hline
\end{tabular}
\end{table}

\section{Conclusion}
In this work, we define the task of detecting structural infringement in both authentic artistic images and those generated by diffusion models. The challenge of this task lies in the lack of training data. We introduce a novel data synthesis pipeline to create image pairs with high structural similarity and low semantic similarity. Using this synthesized data, we extract image structural representations to effectively detect structural infringement. Additionally, we develop two test sets to evaluate detection capabilities. Detecting structural infringement is crucial for protecting creators' rights and advancing AIGC technologies. We hope this work provides valuable insights and inspiration for related fields.

\bibliographystyle{ieeetr}
\bibliography{main}
\end{document}